\documentclass{article}
\usepackage{kr}

\usepackage{times}
\usepackage{soul}
\usepackage{url}
\usepackage[hidelinks]{hyperref}
\usepackage[utf8]{inputenc}
\usepackage[small]{caption}
\usepackage{graphicx}
\usepackage{amsmath}
\usepackage{amsthm}
\usepackage{booktabs}
\usepackage{algorithm}
\usepackage{algorithmic}
\usepackage{xcolor}
\usepackage{framed}
\usepackage{ragged2e}

\usepackage[all]{hypcap} 
\hypersetup{
    unicode=false,          
    pdftoolbar=true,        
    pdfmenubar=true,        
    pdffitwindow=false,     
    pdfstartview={FitH},    
    pdftitle={Class Introspection: A Novel Technique for Detecting Unlabeled Subclasses by Leveraging Classifier Explainability Methods},    
    pdfauthor={Patrick Kage, Pavlos Andreadis},     
    pdfcreator={Patrick Kage},   
    pdfproducer={Patrick Kage}, 
    pdfkeywords={explainability}{latent structure}{classification}{class introspection}, 
    pdfnewwindow=true,      
}

\graphicspath{ {./img/} }
\newboolean{draftenabled}
\setboolean{draftenabled}{false} 

\ifthenelse{\boolean{draftenabled}}
{
    \newenvironment{draft}{
       
       \MakeFramed{\advance\hsize-\width\FrameRestore}}
     {\endMakeFramed}

    \newcommand{\draftinline}[1]{\colorbox{pink}{#1}}
}{
    \newcommand{\draftinline}[1]{}
    \newenvironment{draft}{
       
       \MakeFramed{\advance\hsize-\width\FrameRestore}}
     {\endMakeFramed}
}

\newcommand{\bold}[1]{\textbf{#1}}

\title{Class Introspection: A Novel Technique for Detecting Unlabeled Subclasses by Leveraging Classifier Explainability Methods}

\author{%
Patrick Kage$^1$\and
Pavlos Andreadis$^1$ \\
\affiliations
$^1$University of Edinburgh School of Informatics
\emails
\{p.kage, pavlos.andreadis\}@ed.ac.uk
}

\begin{document}

\maketitle

\begin{abstract}
    Detecting latent structure within a dataset is a crucial step in performing
    analysis of a dataset. However, existing state-of-the-art techniques for
    subclass discovery are limited: either they are limited to detecting very
    small numbers of outliers or they lack the statistical power to deal with
    complex data such as image or audio. This paper proposes a solution to this
    subclass discovery problem: by leveraging instance explanation methods, an
    existing classifier can be extended to detect latent classes via
    differences in the classifier's internal decisions about each instance.
    This works not only with simple classification techniques but also with
    deep neural networks, allowing for a powerful and flexible approach to
    detecting latent structure within datasets. Effectively, this represents a
    projection of the dataset into the classifier's ``explanation space,'' and
    preliminary results show that this technique outperforms the baseline for
    the detection of latent classes even with limited processing.  This paper
    also contains a pipeline for analyzing classifiers automatically, and a web
    application for interactively exploring the results from this technique.
\end{abstract}

\section{Introduction}

\subsection{Motivation}\label{section:intro-motivation}

Training classifiers for machine learning tasks requires that the data is
accurately and completely labeled for a specific application. However, in the
real world there is often more structure to the data than is labeled---and this
can have real-world consequences to how the model performs in production
processes. Within each labeled class, there can be significant variations that
the model picks up on but is invisible to the user---this is \textit{latent
structure} within the class. For an example of this latent structure, consider
a hypothetical classifier to determine whether an image contains apples or
oranges. Oranges tend to be uniformly orange, but apples can come in more than
one color: red, green, yellow, etc. If the labels for the dataset are just
\texttt{apple} and \texttt{orange}, then the information about the color of the
apples is lost. The intuition here is simple: the classifier may know about the
color difference between two types of apples, but still label both apples due
to the training data available to it. Therefore, by using explainability
techniques, a human can detect the different unlabeled subclasses by analyzing
\textit{how} the classifier determines the class of an instance.

As a less trivial but more impactful example, clinical trial results
interpreting the efficacy of a drug's treatment via a classifier may not fully
capture all the subgroups in the input. Are the reasons Person A and Person B
respond to a specific round of treatment similar? How about why Person C did
not respond? This may be down to some specific structure in the input data
which may not be fully captured by training data labeling.


With current methods, it is difficult to determine this latent
structure---especially with high-dimensional or complex data such as image,
audio, or video inputs. Current state-of-the-art methods generally require
either that an entirely new representation is trained (as in the case of
mixture models \cite{bell_automatic_2021} which must relearn the data
distribution), that instances be categorized manually (as in subgroup analysis
\cite{lanza_latent_2013}), or they are only suited to discovering individual
anomalous instances (as in commonality metrics \cite{paterson_detection_2019}).
These methods are discussed more fully in
Section~\ref{section:background-latent-structure}, but in general they are not
one-size-fits-all, and require a separate (new) model to detect latent
structure. Instead, class introspection allows for the re-use of an existing
classifier allowing for analysis to inherit the statistical power of the
classifier.

\subsection{Objectives}


This project aims to provide a solution for the latent structure problem by
leveraging explainability techniques to detect latent (unlabeled) subclasses in
the input data, using a novel approach dubbed \textbf{class introspection}.
This technique compares a classifiers decision making process for each point in
a dataset, and within each predicted class performs clustering over the local
model explanations. Crucially, this \textit{re-uses the existing classifier},
and does not require a separate model for detecting latent structure (in
contrast to the current state-of-the-art). At a high level, the
decision-making process of a classifier model will be different for different
inputs---leading to clusters corresponding to fragmentation within input
classes. This provides a level of auditability on the model training process,
both by ensuring that models are producing results for the correct reasons and
by allowing for the detection of deficiencies in the model setup priors; by
detecting latent structure, possible errors in labeling can be detected in
model training. Additionally, this technique is agnostic to both the
architecture of the particular classifier model used and the particular local
explanation method, allowing for use in even black-box environments (where no
knowledge of the classifier's internals is required).

\section{Background}\label{chapter:background}

\begin{draft}
    Todo: reference
\end{draft}

\subsection{Overview}

In this section, we will discuss the algorithms and methods that are
fundamental to class introspection (namely, explainability and clustering
methods). Additionally, this section discusses several existing algorithms for
detecting structure in data and their limitations as compared to class
introspection.

\subsection{Explainability}\label{section:intro-explainability}

A key issue facing machine learning is explain\-ability (XAI). In current
state-of-the-art machine learning models, the model outputs are generated
opaquely---that is, the reasons that the model chooses to assign one label
rather than another are inscrutable from the outside. While this may be fine in
trivial applications such as face detection, in safety-critical applications the
reasoning for why a model produces the outputs it does can be a literal case of
life and death. Explainability allows for a model to be audited, and for faulty
behaviors to be explained and calibrated away \cite{ghai_explainable_2020}. Even
more crucially, explainability allows for a model's decisions to be trusted by
other agents (like humans) \cite{dosilovic_explainable_2018}.

\subsubsection{What is an explanation?}

Explanations are, at their core, simply an indication of which input features
are relevant for a specific output from a network, and for each how strongly
(or not) these features contributed to the output. Typically this is
represented numerically, where each input feature is weighted by its importance
to the overall classification with respect to the other
features \cite{dosilovic_explainable_2018}.

\begin{figure}[!t]
    \centering
    \includegraphics[width=\linewidth]{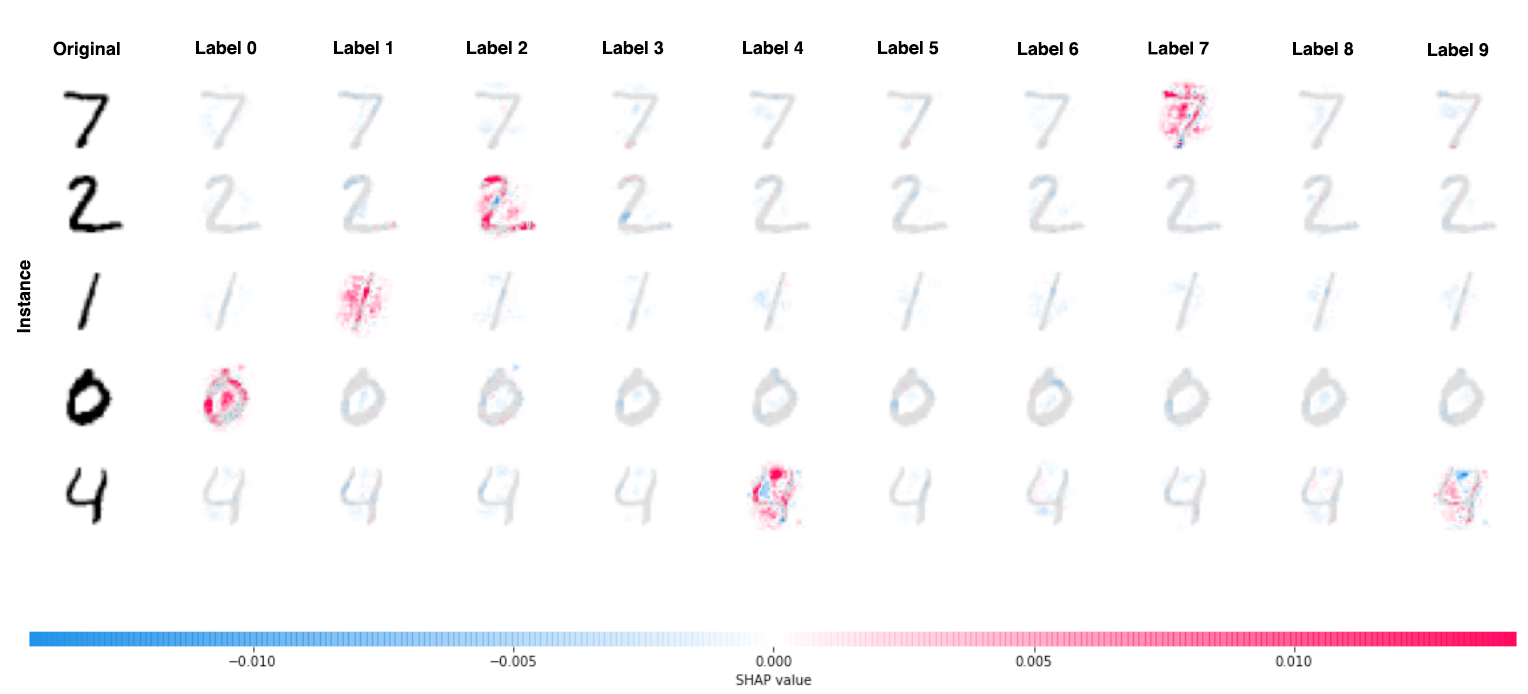}
    \caption{Example of a series of explanations of image data. Here, regions
    of the image that contribute positively towards the classification are red,
    and similarly blue regions denote negative contributions. Note that SHAP
    creates explanations for each instance for each label, regardless of true
    label. For more information, see
    Section~\ref{section:intro-neural-nets}.}\label{fig:intro-shap-explanation}
\end{figure}

In tabular data, numerically weighting input features trivially shows which
inputs are important as the feature is clearly defined (i.e. each input feature is labeled). Image data is more complex, as the
features do not carry the same amount of information per-feature as tabular
data; explanations for this type of data show which regions of the image were
important to a model's classification (see
Figure~\ref{fig:intro-shap-explanation}).

\subsubsection{Inherently Interpretable Models}

The easiest way to generate explanations for models is by using models that are
inherently interpretable. There are two common model architectures with this
feature: regression models and decision trees. Decision tree outputs can be
simply explained by following the chain of decisions from the root node down to
the eventual leaf representing the classification. Regression models are
similarly simple, with linear regression and logistic regression:

\[
    \text{linreg}(\bold x; \bold w, b) = \bold w^\top \bold x + b
    \qquad
    \text{logreg}(\bold x; \bold w, b) = \sigma(\bold w^\top \bold x + b)
\]

We can see that for each case, there is a one-to-one correspondence between the
weight vector \(\bold w\) and the input vector \(\bold x\). Therefore, the
relative importance of each input feature in \(\bold x\) is directly encoded in
the weights of the model---exactly an explanation.

Of course, using regression models or decision trees may not be an optimal
strategy as these models are very limited in their capabilities. Another option
is to converting an existing model as a whole into a more easily interpretable
model architecture. For example, an interpretable model can be created with
\textit{Self-Explanatory Neural Networks} (SENNs) which are an extension of
logistic regression \cite{teso_toward_2019}. This can use very few weights, and
it is easy to determine the reasoning for any given classification as the
weights correspond to a positive or negative linear combination of inputs. This
approach, however, is limited in that some algorithms are not well-suited to
interpretability. Algorithms such as deep neural networks are uninterpretable,
and their performance cannot easily be matched by interpretable models.

\subsubsection{Local Outcome Modeling}

If the model to be explained cannot be converted to an easily-interpretable
model, it is still possible to create an explanation of its behavior using
black-box explanation methods.

\begin{figure}[!t]
    \centering
    \includegraphics[width=\linewidth]{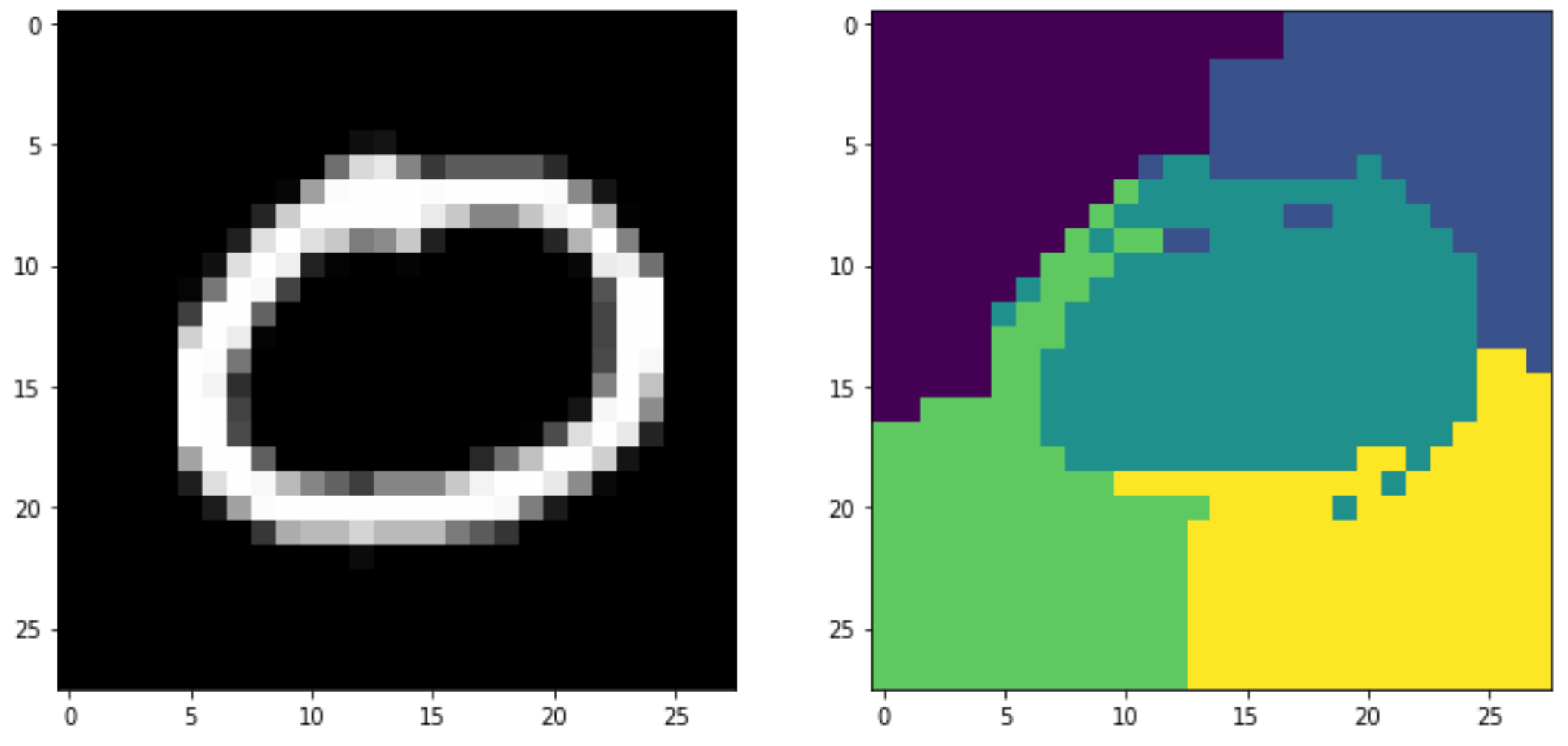}
    \caption{Example of LOO segmentation.}\label{fig:intro-loo-segmentation}
\end{figure}

The simplest black-box approach is the Leave-One-Out (LOO)
algorithm \cite{abdalla_visual_2021}. LOO simply segments the input features,
and for each segment zeros out the segment and runs the model inference to
determine how much the output changes. Despite LOO's apparent simplicity, this
strategy is surprisingly effective at determining the salient regions of an
input, and being a black-box algorithm means it can be broadly applicable to a
diverse variety of algorithms \cite{abdalla_visual_2021}. Careful selection of
segmentation algorithms can lead to very accurate saliency maps (see
Figure~\ref{fig:intro-loo-segmentation}). However, this flexibility comes at a
cost.  LOO is expensive to compute, as each segment requires a re-inference;
additionally, this method does not take into account inter-dependencies between
regions \cite{abdalla_visual_2021}. Additionally, this does not give per-feature
saliency mapping, rather operating over superpixels---useful for individual
explanations, but less so for comparing explanations between instances.

A more robust approach is to train a smaller model to explain individual
predictions of a dataset \cite{teso_why_2018}. This local approach has the
benefit that the overarching model can be a black box while still providing
explanations of the model behavior. This works because even if the decision
surface may be defined by many uninterpretable features, a single point in that
decision surface can be locally approximated. The most popular implementation
of this approach is Local Interpretable Model-agnostic Explanations
(LIME) \cite{ribeiro_why_2016}. LIME fits a flat plane against the decision
surface, and infers the boundary by perturbing the input point to generate a
cloud of points which are then all inferred by the model. While not being
perfectly accurate, this is effective at broadly showing the explanation of the
point inference and has the advantage of being model-agnostic. LIME still
suffers from some the same limitations of LOO:\ each explanation requires
multiple inferences (due to the input point perturbation), and to work over
large dimensional data (such as images) superpixel segmentation must be used.

\begin{figure}[!t]
    \centering
    \includegraphics[width=8cm]{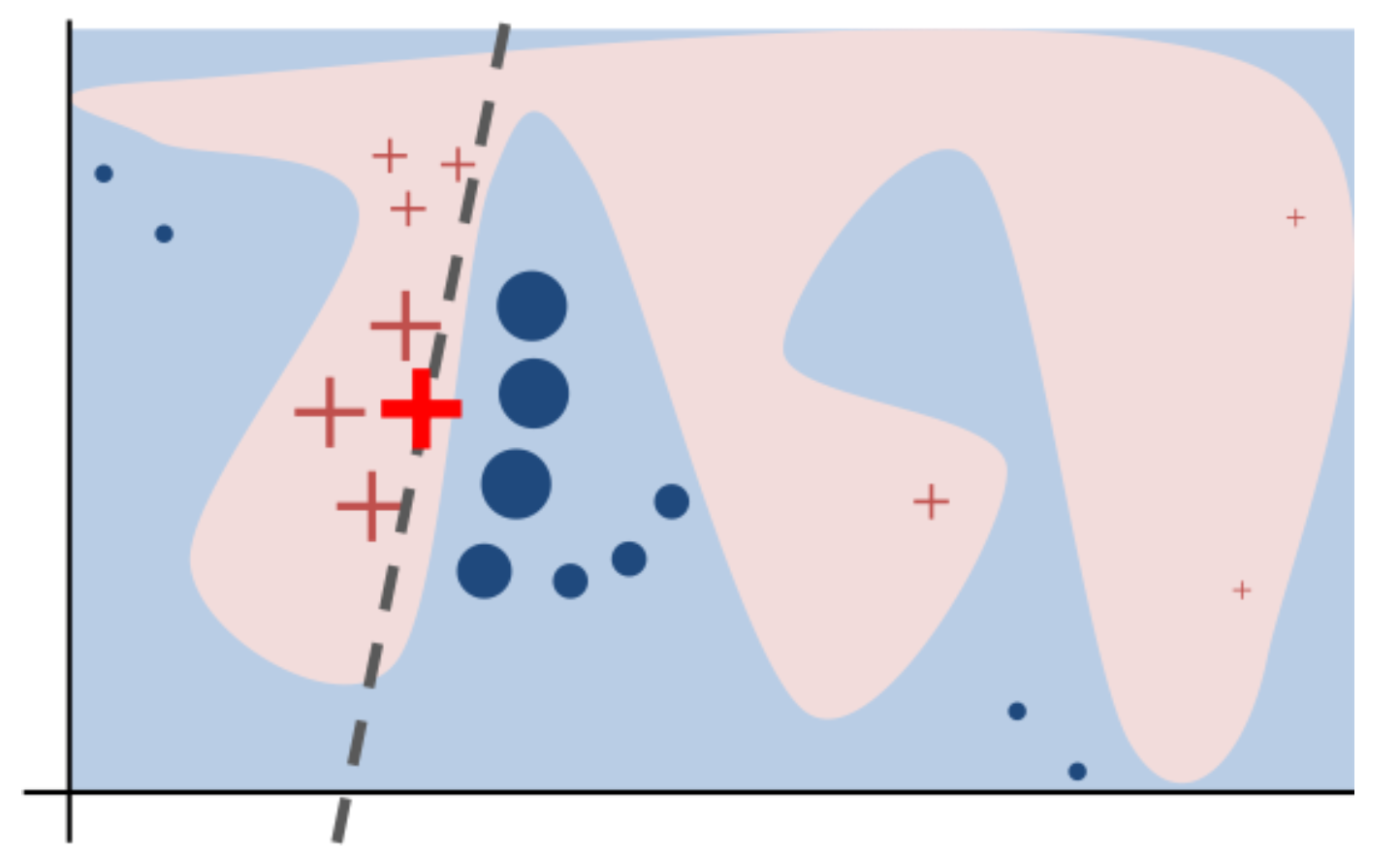}
    \caption{LIME local
    model. From Ribeiro et.\ al.\protect \cite{ribeiro_why_2016}.}\label{fig:intro-lime-fitting}
\end{figure}

\subsubsection{Neural Network Approaches}\label{section:intro-neural-nets}

A final approach is specific to neural networks: a series of techniques are
available to compute the gradient of the class score with respect to the input
pixels, yielding an effective explanation of which inputs are salient to the
model's ultimate classification \cite{abdalla_visual_2021}. These methods are
typically white-box, and operate over the specific structure of the neural
network; this means that explanation method implementations need to be aware of
the implementation details of the networks they are explaining.

\begin{figure}[!t]
    \centering
    \includegraphics[width=8cm]{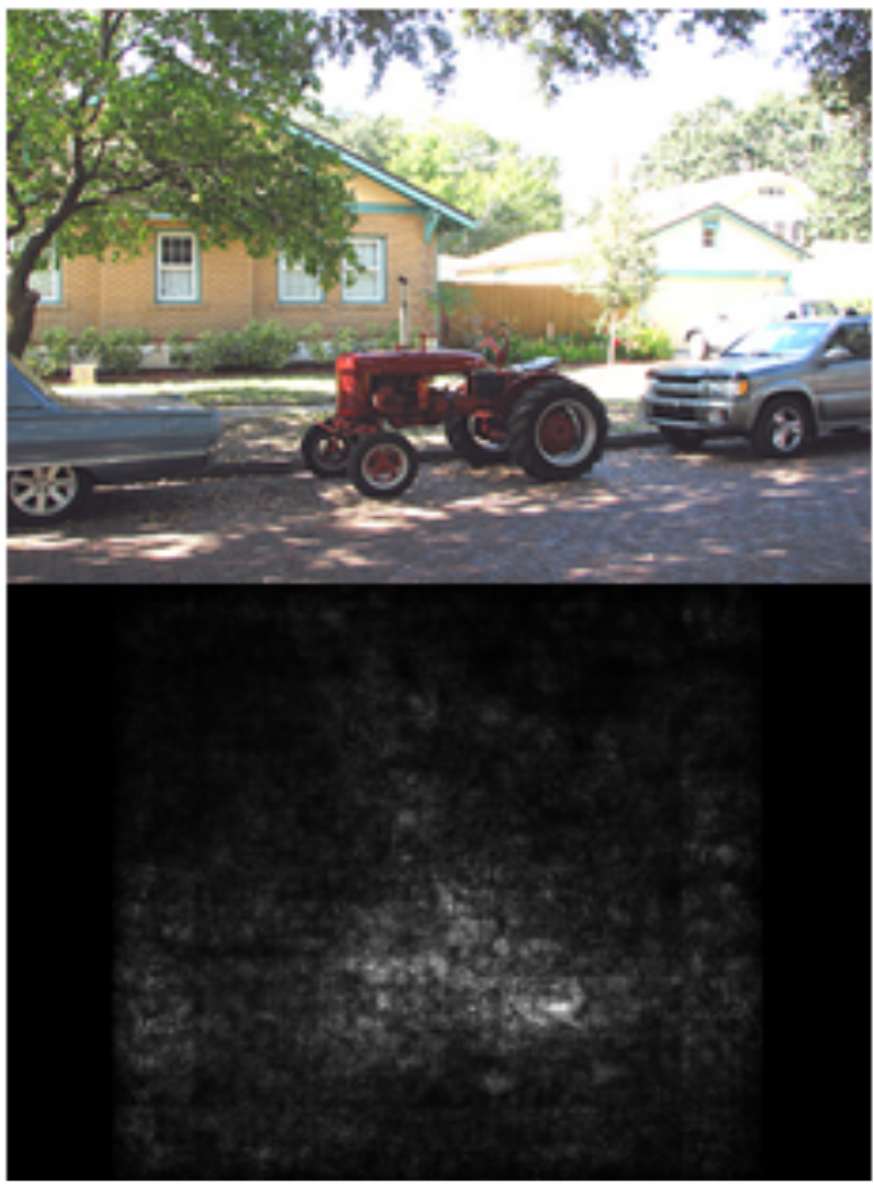}
    \caption{Example saliency map generated with gradient ascent, depicting the
    saliency map from the top class prediction from a ConvNet~\protect \cite{simonyan_deep_2014}.}\label{fig:intro-gradient-ascent-saliency}
\end{figure}

There are a several methods of achieving this, with the simplest being Gradient
Ascent \cite{simonyan_deep_2014}. In gradient ascent, the gradient of the class
label with respect to each input feature is computed via backpropagation all in
one go, leading to a per-feature importance score across the image
\cite{abdalla_visual_2021}. However, this can cause a very grainy saliency map
(see Figure~\ref{fig:intro-gradient-ascent-saliency}). A more nuanced approach
is Deep Learning Important FeaTures (DeepLIFT), which compares the activations
of network neurons to a ``reference activation'' (activations on a different
instance) and can separate positive and negative contributions to give
higher-quality results than Gradient Ascent
\cite{abdalla_visual_2021,shrikumar_learning_2019}.

A popular library for generating these explanations is Shapley Additive
Explanations (SHAP) \cite{lundberg_unified_2017}. SHAP represents explanations
as a set of Shapley values (a measure of how important a contribution is to an
overall whole borrowed from game theory) of the overall model, and computes
these using a unification of several techniques (notably LIME and
DeepLIFT) \cite{lundberg_unified_2017}. This approach generates high-quality
explanations and is packaged into an easy-to-use Python library
\texttt{shap} \cite{slundberg_shap_2021}.

SHAP is used heavily throughout this project, so it is worth examining its
underlying principles of operation. SHAP generates explanations for each label
for each instance; for example for each numeral in MNIST SHAP generates ten
sets of explanations: one for label \texttt{0}, one for label \texttt{1}, and
so on. This is visible in Figure~\ref{fig:intro-shap-explanation}, where each
instance (rows) has ten different Shapley value sets (columns). These Shapley
values are not necessarily the same magnitude between data points (though in
this project they usually are), and are typically similar within a dataset.
SHAP operates over not only image data (with DeepSHAP) but works with other
types of data, such as tabular data.

\subsection{Hierarchical Clustering}




Hierarchical clustering algorithms are a series of algorithms to form clusters
over data where the number of target clusters is not necessarily known;
contrasting against partitioning algorithms (such as K-means clustering) which
require a knowledge of the cluster count to partition the dataset. Hierarchical
clustering algorithms work by combining data points into clusters
agglomeratively, and this property makes them incredibly powerful for analyzing
unknown data as they are able to discover the
data \cite{ester_density-based_1996}.

A commonly-used clustering technique is Density Based Spatial Clustering of
Applications with Noise (DBSCAN). DBSCAN is able to efficiently handle
clustering over a dataset with arbitrary-shaped clusters, and is able to
achieve this with minimal required knowledge of the underlying dataset (as it
is parameterized by a single hyperparameter) \cite{ester_density-based_1996}.
This is achieved by picking an arbitrary point in the dataset and building a
list of points reachable within a distance \(\epsilon\) from that point
(measured via a Euclidean distance metric). If that list is over a threshold
number of points (typically 5), the list of points is given a cluster
label---otherwise, it is marked as noise. This process is repeated until all
points are either assigned a cluster or marked as noise.

However, due to DBSCAN's reliance on a Euclidean distance metrics, it is not
suitable to very high-dimensional data. This is due to \textit{curse of
dimensionality}, which states that as the dimensionality goes up the more
sparse the input space becomes for nearest-neighbor
searching \cite{marimont_nearest_1979}.

\subsection{Latent Structure
Detection}\label{section:background-latent-structure}



Class introspection is a novel technique, but the idea of identifying latent
classes within a dataset is not new. A prominent application area of this
concept is medical studies, where similarities between subgroups of patients
need to be identified in order to determine the safety of a drug; for example,
if certain members of the experimental group had different reactions to a
research pharmaceutical, then it would be important to both identify
fragmentation inside that group (i.e.\ had a reaction to a drug in different
ways) and to identify common features between those individuals (e.g.\ similar
ages, etc.) \cite{lanza_latent_2013}. Discovering these similarities is crucial,
as treatment techniques can rely on a specific confounding variable in a
population that is not readily captured by the available features.

\subsubsection{Subgroup Analysis}

Subgroup analysis is a simple method of detecting latent structure. Subgroup
analyses are performed by manually categorizing gathered data into subgroups
based on some common characteristic of the data, and are typically examined by
incorporating some moderating variable into a regression and interpreting the
results \cite{higgins_cochrane_2011}. However, because this grouping is
inherently observational, this method is difficult to use effectively and is
time-intensive. Additionally, the results of such analyses are subject to high
Type I errors, as false positives are easy to make when manually grouping
data \cite{lanza_latent_2013}. 

\subsubsection{Finite Mixture Models}

Finite mixture models are another way of determining latent structure (or
structure in general) \cite{lanza_latent_2013}. Finite mixture models represent
the data as a linear combination of component densities and maximize the
likelihood of a specific configuration of models, and a common density function
to use is the Gaussian normal distribution \cite{bell_automatic_2021}. Gaussian
mixture models optimize the probability:

\[
    p(x) = \sum^M_{m=1} P(m) \mathcal{N} (x; \mathbf{\mu}_m, \Sigma_m) 
\]

where the probability \(p(x)\) of a specific point is given by a sum of normal
distributions moderated by mixing parameters
\(P(m)\) \cite{bell_automatic_2021}. This probability is typically optimized by
the expectation-maximization (EM) algorithm \cite{siedel_mixture_2011}.

Finite mixture models can find latent structure, but they are limited in their
effectiveness for this task. Finite mixture models are limited to discovering
structure in the form of the basis functions chosen (e.g. Gaussians).
Additionally, mixture models must re-learn the data distribution without any
guidance from the original class labels which is computationally
expensive\cite{bell_automatic_2021}. FMMs also generalize poorly to
high-dimensional data, as it is difficult to fit models in high-dimensional
spaces. In general, these techniques are poorly suited to finding latent
structure in complex data.

\subsubsection{Commonality metrics}\label{subsection:intro-commonality-metrics}

A cutting edge technique in this area is the detection of rare subclasses via
\textit{commonality metrics} \cite{paterson_detection_2019}. This technique
analyzes the input training classes for a given dataset and for each class
determines the average activation in the penultimate neural layer for that
class, and then scores each instance based on how similar it is to that average
activation.

This is technique does not identify latent classes in and of themselves;
rather, it identifies single instances that may be mislabeled or are
significantly different than the average. Additionally, because the commonality
metric approach operates over an average activation, large numbers of
far-from-average instances will skew the whole commonality metric---potentially
rendering the method less effective at identifying singular anomalous
instances.

\begin{draft}
    Under construction: papers for referrer \#1 comments. See
    \cite{asano2020selflabelling} and \cite{caron2019deep}.
\end{draft}

\subsubsection{Clustering over neural networks}

Additionally, several recent techniques directly address the problem of
unsupervised labeling of datasets. Broadly, these techniques introduce a
clustering step during the training of a neural network to improve the labeling
performance of the network. Two such techniques are DeepCluster
\cite{caron2019deep}, and the self-labeling method proposed by Asano et. al.
\cite{asano2020selflabelling}, where the loss function jointly learns the
neural network parameters and the cluster assignments of the inferred features.
DeepCluster itself simply uses \(k\)-means clustering to achieve this, and
Asano et. al. use a more complex linear-programming-based method. Both,
however, significantly outperform other feature-based learning approaches
\cite{caron2019deep,asano2020selflabelling}.

\section{Methodology}

\subsection{Problem Formulation}


The central problem to be solved is the discovery of latent subclasses in
the input space. The aim is to find unlabeled (latent) subclasses within
labeled classes in an existing dataset, with the hypothesis that these latent
subclasses can carry additional information that is relevant to the authors of
the classifier (see the examples in Section~\ref{section:intro-motivation}).
The latent subclasses should be differentiated from one another without human
intervention; that is to say, this should be an unsupervised technique.

This is, at its core, a clustering problem. In the ideal case (with no latent
structure), each discovered cluster should correspond to a single true label in
the dataset. However, in the case where latent subclasses exist within a class,
we expect to see two or more cluster labels assigned to a single class label.
In general, this method should take a dataset as input and the labels and
output a list of cluster labels for each instance. As a baseline technique,
simple heirarchical clustering could be performed; however, by using the
explainability methods the resulting projection is hypothesized to be easier to
cluster over.

\subsection{Method Overview}

Class introspection's central hypothesis is that even if a instances within a
dataset is all labeled identically, the specifics of how a specific classifier
\textit{interprets} that instance can change significantly between those
instances---and it is precisely that change that can be used to determine the
existence of latent subclasses. In other words, this approach solves the
subclass detection problem by leveraging an existing classifier, and through
that the classifier's statistical power. Detecting these instance changes
relies on the combination of several methods discussed in
Chapter~\ref{chapter:background}. The general process is described below:

\begin{enumerate}
    \item Generate classifications for all instances in the dataset.
    \item For each instance, create a local explanation of the classifier's
        decisions.
    \item For each class:
        \begin{enumerate}
            \item Select all instances in the dataset classified as the
                selected class, and their explanations.
            \item Run a clustering algorithm over the class's explanation.
            \item \textbf{Multiple clusters indicate the existence of latent
                structure.}
        \end{enumerate}
\end{enumerate}

\begin{figure*}[ht]
    \centering
    \includegraphics[width=\linewidth]{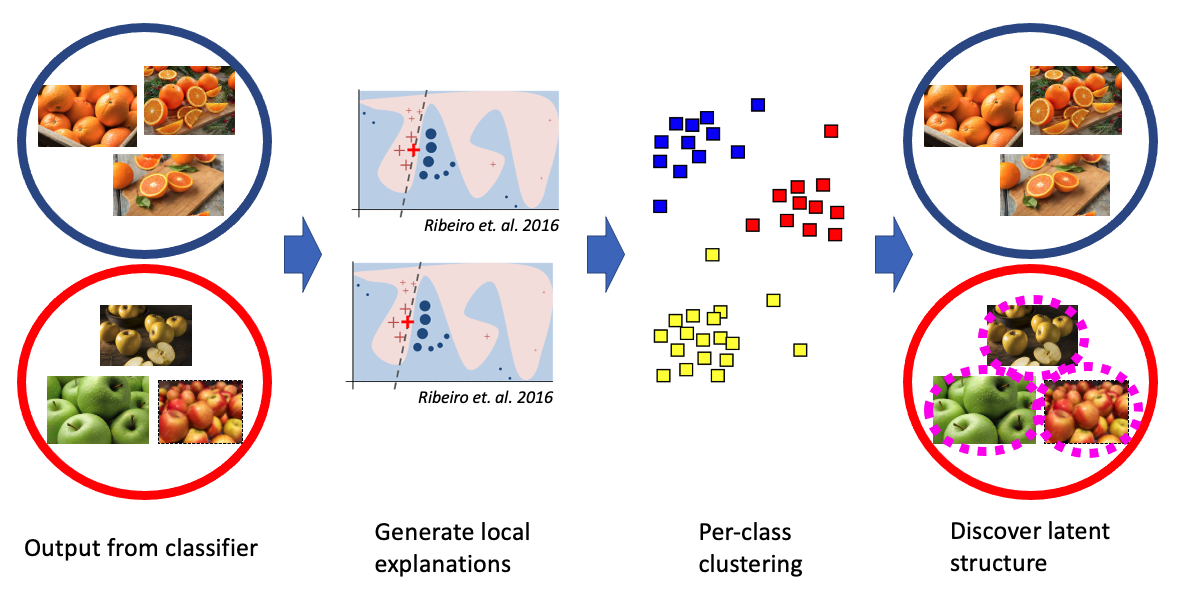}
    \caption{Class introspection methodology
    overview.}\label{fig:meth-methodology-overview}
\end{figure*}

Clustering over the instance explanations is possible as these explanations act
as a \textit{proxy} for the underlying instance. The explainability methods
discussed in Section~\ref{section:intro-explainability} produce saliency maps
over the input features \cite{ribeiro_why_2016,lundberg_unified_2017}, and these
saliencies are expressed as a tensor of saliency weights over the initial
features. This is crucial: because these saliency maps are numeric, they can be
interpreted just as easily as the original data---or even more easily, as
discussed further on in this paper.

Because the explanations are simply numeric they can be clustered over, but the
clustering itself is a challenge. Because the total class count is unknown,
partitioning algorithms (e.g. K-means clustering) are not suitable to the
task---if the total class count was known, the problem would be solved as we'd
already know which subclasses exist. Instead, agglomerative clustering methods
are appropriate for this use case, as they can operate over an unknown number of
clusters. In this paper, DBSCAN is used to apportion the explanations into
clusters---although others can be used (see Section~\ref{section:future-work}).

Before clustering can be applied, the dimensionality of the data must be
reduced. Clustering algorithms rely on (typically Euclidean) distance metrics,
and as the number of dimensions increases the number of unit cells in that
space increases exponentially, and so in these high dimensional spaces the
nearest point can be extremely far away \cite{scarpa_data_2011}. In this paper,
principal component analysis is used to reduce the number of dimensions before
clustering, yielding much better results.

At the end of the clustering process, the assigned cluster labels within each
class are displayed as a histogram. The key insight here is that the cluster
labels correspond to the similarity of the input images to each other, and if
there are a large number of instances in two or more clusters those high-count
are a candidate for a latent subclass.

Ultimately, this method requires human interpretation of the results as some of
the detected latent structure may be intentional (e.g.\ in the apples-oranges
example the classifier may not care about the different kinds of apple). Even
so, the output from this algorithm is much more readily interpretable than
attempting to audit the dataset and labels by hand.

\subsection{Baseline}\label{section:meth-baseline}

A simple baseline technique for detecting this latent structure is to ignore
the explanatory methods altogether and perform clustering over the raw instance
data itself. PCA is used to reduce the dimensionality of the data before being
passed to the clustering algorithm, DBSCAN.\@ This approach is similar to the
commonality metrics methods described in
Section~\ref{subsection:intro-commonality-metrics}, and is nearly identical to
the methodology in Figure~\ref{fig:meth-methodology-overview}. To generate a
dataset with latent structure, artificial latent structure is induced as in
Section~\ref{section:artificial-structure}.

\begin{figure}[!t]
    \centering
    \includegraphics[width=\linewidth]{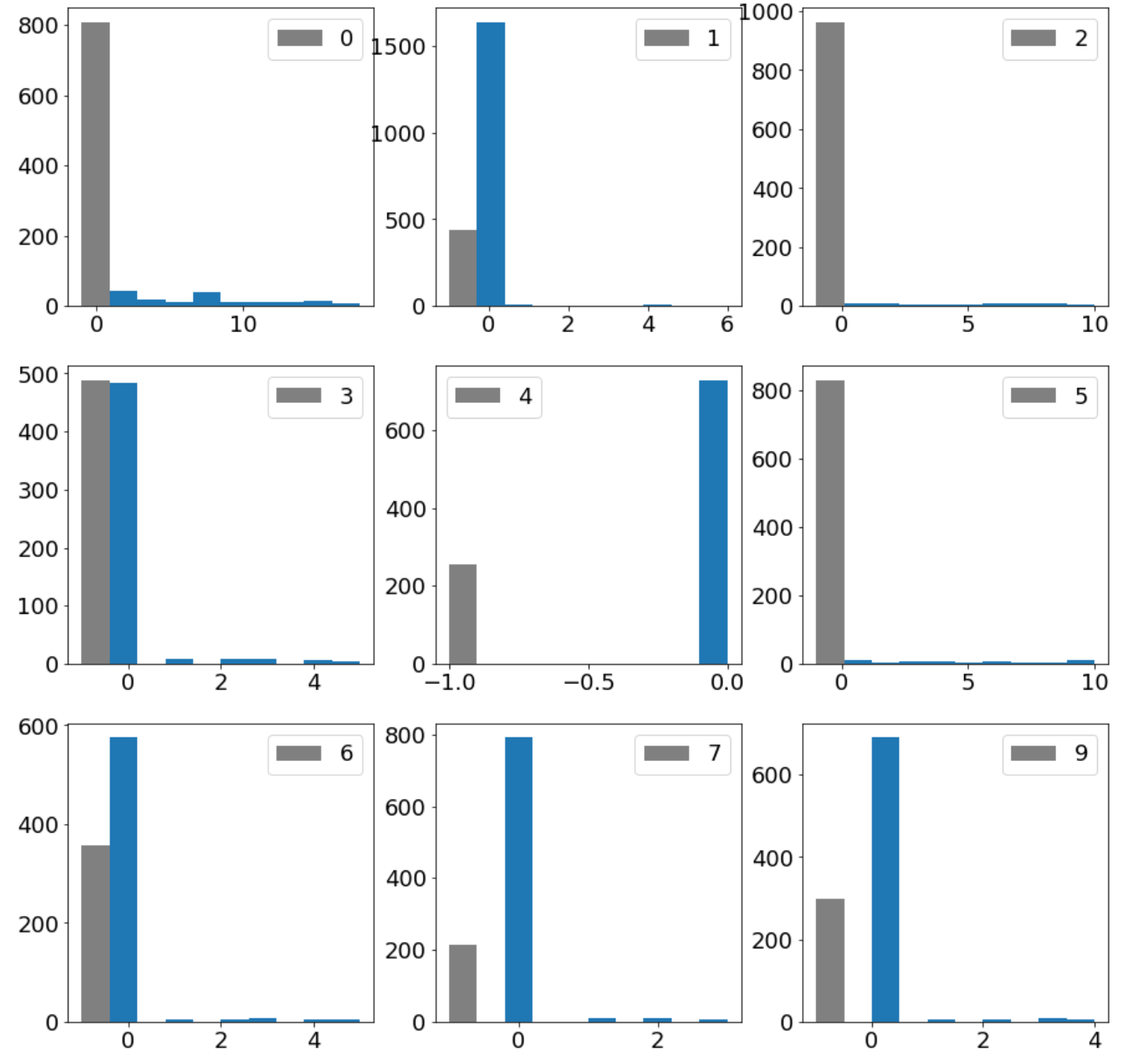}
    \caption{Detected class fragmentation from the baseline run against MNIST
    \(1\to 8\) with \(\epsilon = 250\). Each blue bar represents a detected class from DBSCAN. Note how none of the classes show a
    major split in the assigned cluster labels (ignoring gray noise bars).
    }\label{fig:results-baseline-1-8}
\end{figure}

\begin{figure}[!t]
    \centering
    \includegraphics[width=\linewidth]{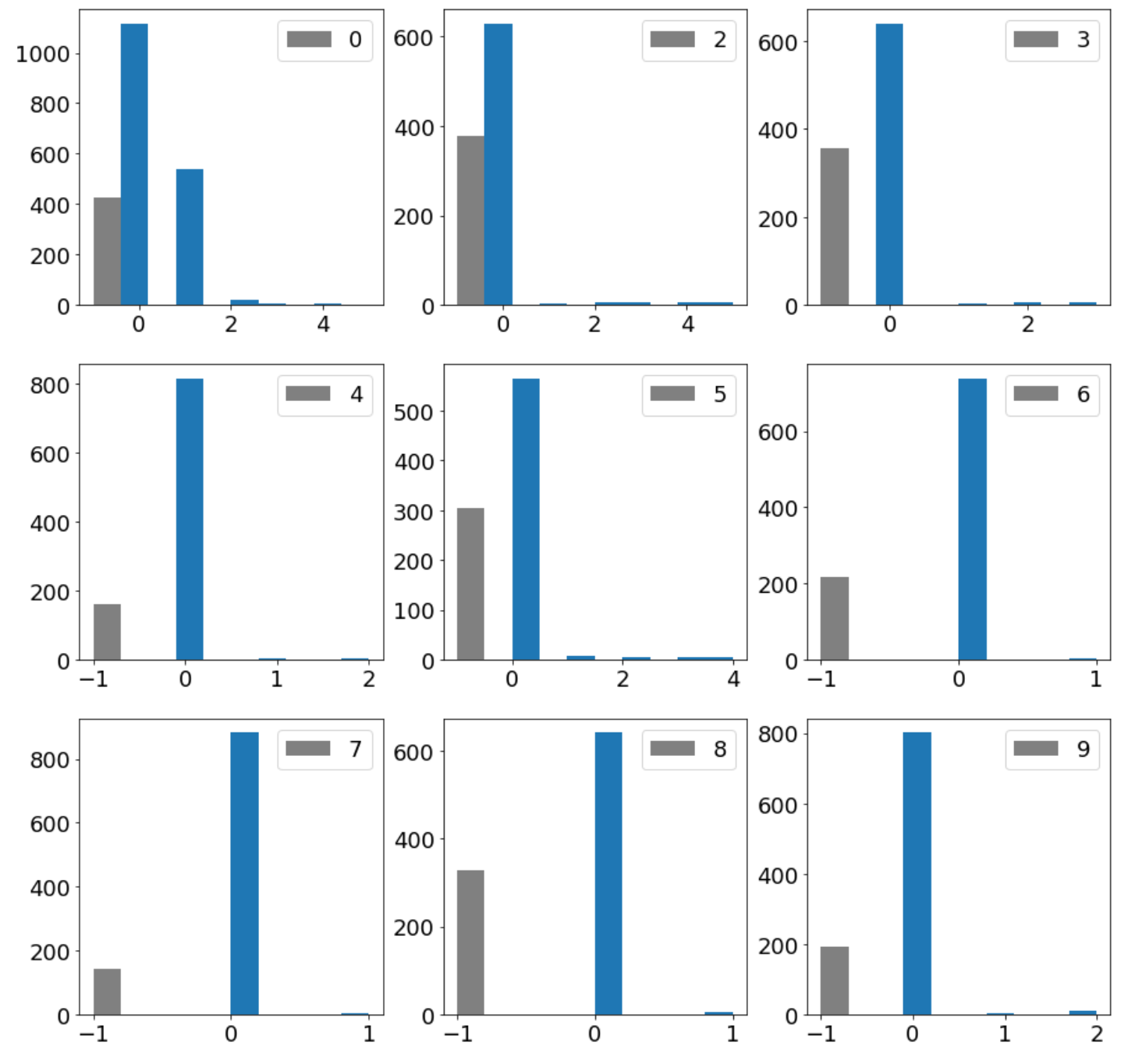}
    \caption{Detected class fragmentation from the baseline run against MNIST
    \(0\to 1\) with \(\epsilon = 275\). Note how here, the latent structure is
    successfully identified (top left).}\label{fig:results-baseline-0-1}
\end{figure}

Running the baseline over the MNIST dataset yields a surprising result: the PCA
and DBSCAN over the raw MNIST digits is not likely to determine when the
artificial latent structure is present. Even with a hyperparameter sweep of the
DBSCAN \(\epsilon\) parameter, in some cases an optimal configuration cannot be
found where the bridged class has a significant split in cluster membership as
opposed to the non-bridged classes. Figure~\ref{fig:results-baseline-1-8}
contains the output for the \((1 \to 8)\) bridging, and it is clear that the
bridged class is not identified: no class contains any split in cluster
membership, and some are entirely noise. However, some splits do work: in the
case of the \(0 \to 1\) split in Figure~\ref{fig:results-baseline-0-1}, the
latent structure is successfully identified.

\section{Experimentation}\label{chapter:initial-explorations}

\subsection{Artificial Latent Structure}\label{section:artificial-structure}

To develop a pipeline to detect latent structure, we first need a dataset which
contains latent structure. Unfortunately, it is difficult to find a dataset
with such structure already in it. The challenge is twofold---either the
dataset is simple enough that latent structure is virtually nonexistent, or the
dataset is complex enough that detecting latent structure requires domain
knowledge to detect. For a useful comparison of class introspection techniques,
the dataset needs both the original class labels and labels for the latent
structure to be detected. Creating these labels requires manually labeling
every instance, which is both incredibly time-consuming and out of the scope of
this project.

\begin{figure}[ht]
    \centering
    \includegraphics[width=\linewidth]{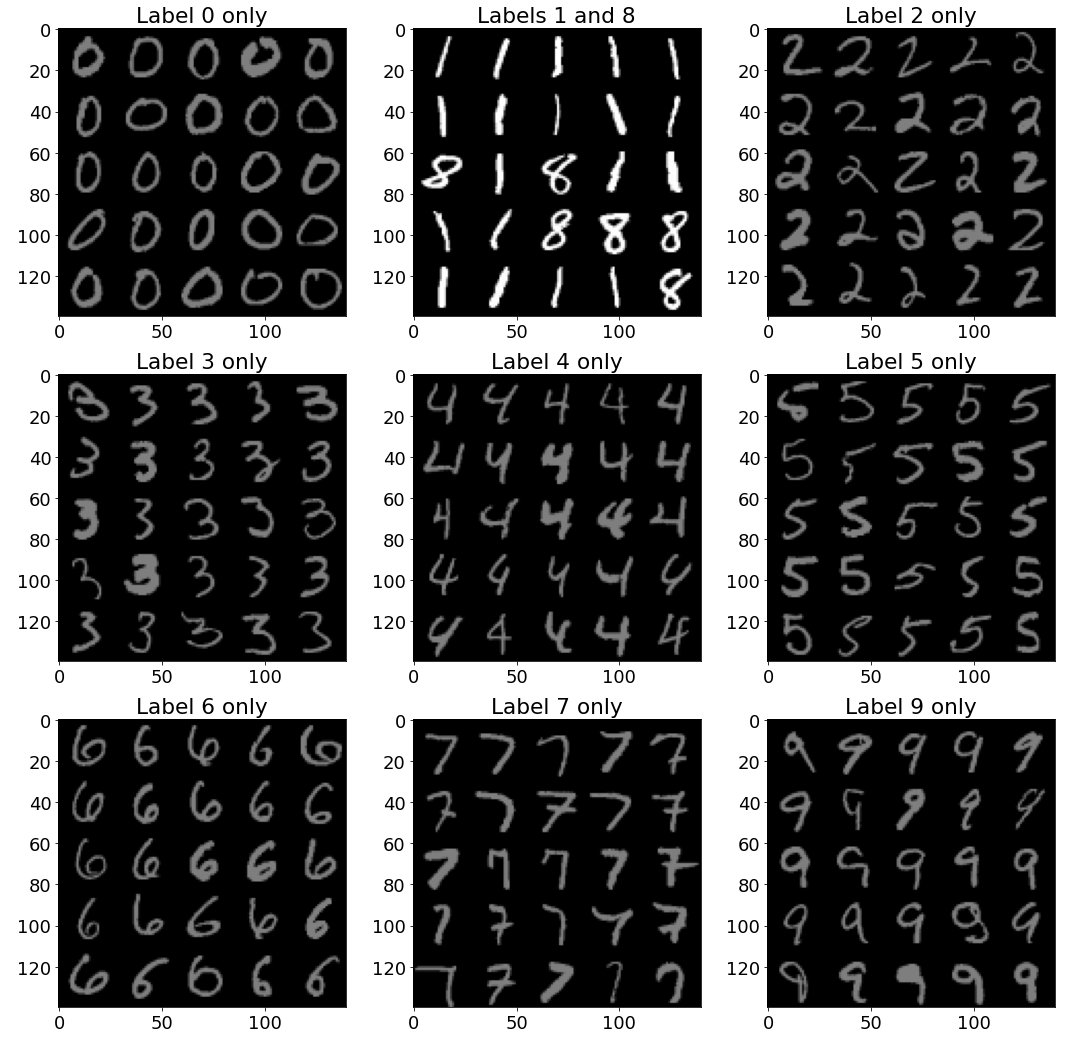}
    \caption{A demonstration of artificial latent structure with MNIST, here bridging the
    \texttt{1} and \texttt{8} classes. Note that the bridged classes have been
    highlighted.}\label{fig:mnist-18-bridge}
\end{figure}

Clearly, another solution is needed. Instead of relying on existing datasets to
have exploitable latent structure, we can \textit{create} latent structure
ourselves by ``bridging'' class labels together. This trivially creates latent
structure, as both the instances for label \(A\) and label \(B\) are bound to
the same class---and thus the new bridged class label has two distinct types of
instance in it. This also has the benefit of giving labels corresponding to the
original classes, as the original labels can be compared to the bridged labels
to show which instances have been bridged. See Figure~\ref{fig:mnist-18-bridge}
for an example with MNIST with labels \texttt{1} and \texttt{8} bridged
together.

This approach has limitations, however. The primary complication is the nature
of the latent structure itself: bridging two classes can create a superclass
with extremely obvious latent structure, which may not align well with
real-life latent structure. However, for the purposes of this project these
effects have been ignored, as it is useful to have obvious latent structure to
detect rather than very subtle structure in order to test the pipeline.
Additionally, bridging class labels can lead to unexpected behavior whilst
training models over a dataset with a small number of classes. This is a
fundamental issue, as bridging two classes effectively removes one of the
classes. For simple datasets, this can cause certain models to not learn the
structure of the superclass but instead simply learn one of the other classes
and segment the input space into ``class and not class''.

\subsection{MNIST dataset}

An appropriate multiclass dataset for the development of the class
introspection algorithm is the MNIST dataset \cite{lecun_mnist_2010} which
contains \(70,000\) handwritten digits (\(0\) through \(9\)) as \(28\) pixel by
\(28\) pixel monochrome images. The MNIST dataset is a common dataset in
machine learning as it is easily understood by beginners while still being
complex enough to allow for meaningful analysis.

For the purposes of class introspection, it is an ideal dataset due to its
relative simplicity and high class count. This high class count is crucial to
avoid the issues with too few meaningful classes discussed above, as with a single
bridged class there are still eight remaining classes.

\subsubsection{SHAP and Keras}\label{subsection:shap-and-keras}

Due to the fine-grained explanations required for class introspection, LIME's
explanations were insufficient. A different explanation engine was required,
and SHAP (with the DeepLIFT backend) is able to provide those explanations.
Instead of operating over superpixels, SHAP uses DeepLIFT to backpropagate the
contributions of every neuron to every feature in the input
space \cite{shrikumar_learning_2019,lundberg_unified_2017}. 

\begin{figure}[!t]
    \centering
    \includegraphics[width=\linewidth]{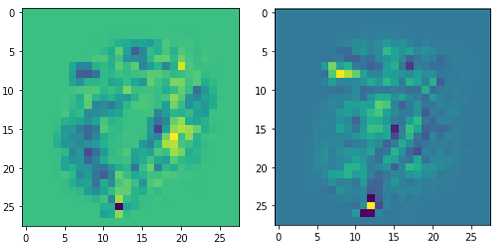}
    \caption{On the left: SHAP explanations over unbridged data, detail on
    explanation for label 0 and true label 7. Note the band of blue values in
    the shape of a zero, these are negative contributions (blue) from the
    features that would match label 0 and are missing on the 7. On the right:
    SHAP explanations over unbridged data, detail on explanation for label 7
    and true label 7. Note the positive contributions (green) along the body of
    the glyph.}\label{fig:mnist-keras-shap-detail-7-0-and-7-7}
\end{figure}

To keep the experiment simple, a relatively simple network architecture was
chosen: fully-connected \(784 \to 128 \to 128 \to 64 \to 10\) layers with ReLUs
and a softmax activation. Training this network resulted in a \(97.91\%\)
accuracy over the test dataset, which is expected over such a relatively simple
dataset. Running SHAP explanations over the test dataset yielded a per-pixel
saliency map, and this map is intuitively interpretably; as an example for an
instance with a true label 7 the explanation for label 0 shows that the network
expects a round shape, and because the instance does not match that shape the
input features negatively contribute towards classification as a 0. For the
same instance, the explanation of the true label 7 shows the shape of the 7
positively influencing the classification of a 7, yielding a classification of
7 for the instance (see Figure~\ref{fig:mnist-keras-shap-detail-7-0-and-7-7}).

\begin{figure}[!t]
    \centering
    \includegraphics[width=\linewidth]{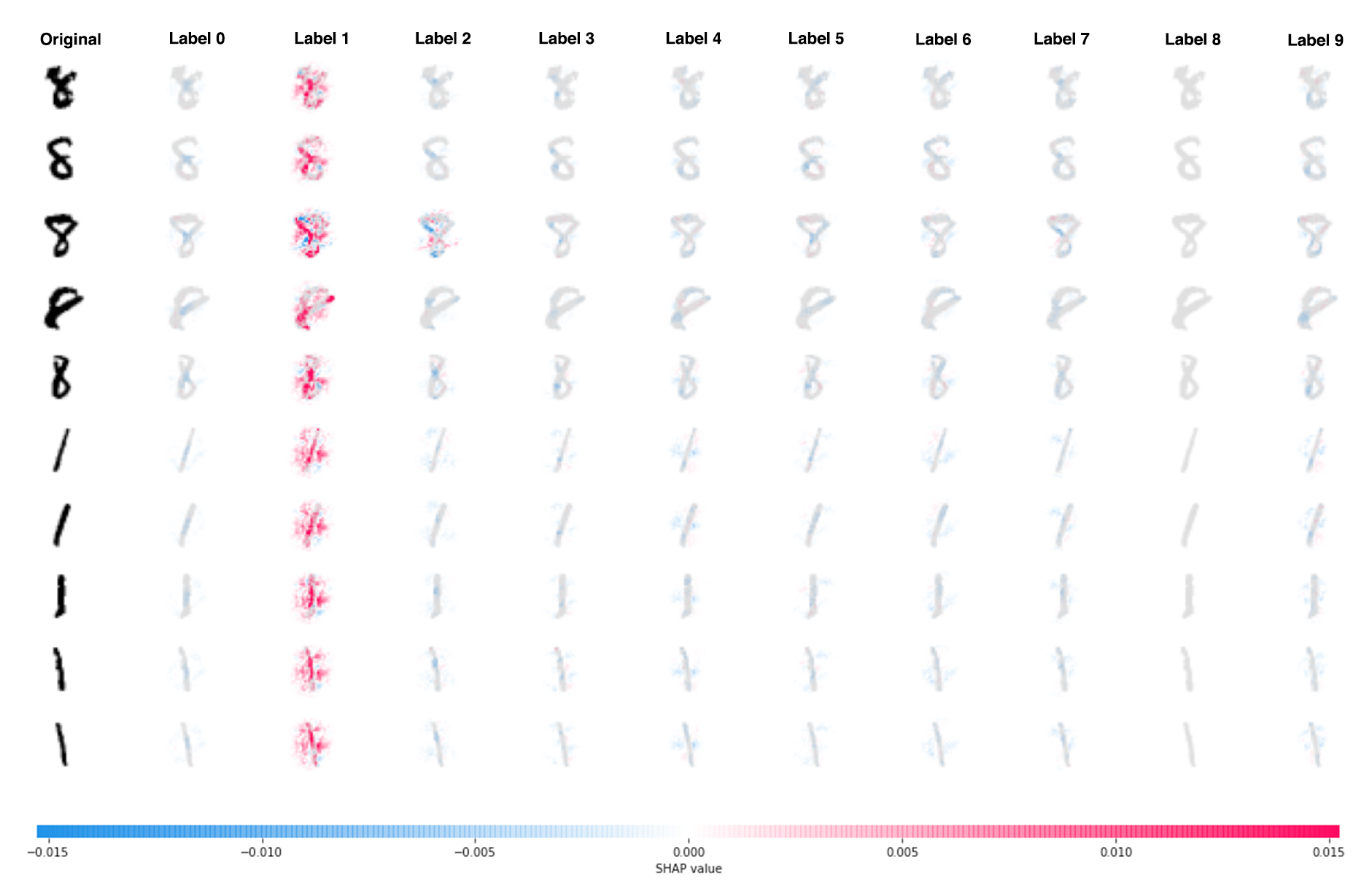}
    \caption{SHAP explanations over bridged MNIST
    data.}\label{fig:mnist-keras-shap-bridged}
\end{figure}

Again, the \(1\) and \(8\) digits were bridged together due to their
dissimilarity, and the network was trained again (achieving an accuracy of
\(97.92\%\)). Figure~\ref{fig:mnist-keras-shap-bridged} shows another grid of
SHAP explanations, note that there are no explanations in the \(8\) column as
they have been bridged with the \(1\) column. Filtering just for instances in
the bridged category shows a clear difference between the explanations for the
\(1\) instances and the \(8\) instances.


The next step in the class introspection pipeline is to cluster the explanation
to isolate the latent structure in the class. This poses several problems: the
number of clusters is unknown, the data may not be linearly separable, and the
data is high-dimensional. To solve the unknown cluster count, hierarchical
clustering is used (DBSCAN). DBSCAN can handle clusters that are not linearly
separable, making it suitable for this
application \cite{ester_density-based_1996}. However, DBSCAN relies on a
euclidean distance metric, requiring dimensionality reduction to avoid the
curse of dimensionality.

\begin{figure}[!t]
    \centering
    \includegraphics[width=\linewidth]{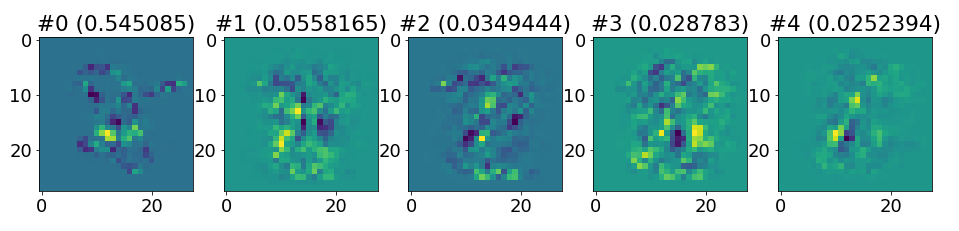}
    \caption{PCA vectors for the \(1\) to \(8\) bridge, with the explained
    variance of each component.}\label{fig:mnist-keras-shap-bridged-pca}
\end{figure}

\begin{figure}[!t]
    \centering
    \includegraphics[width=\linewidth]{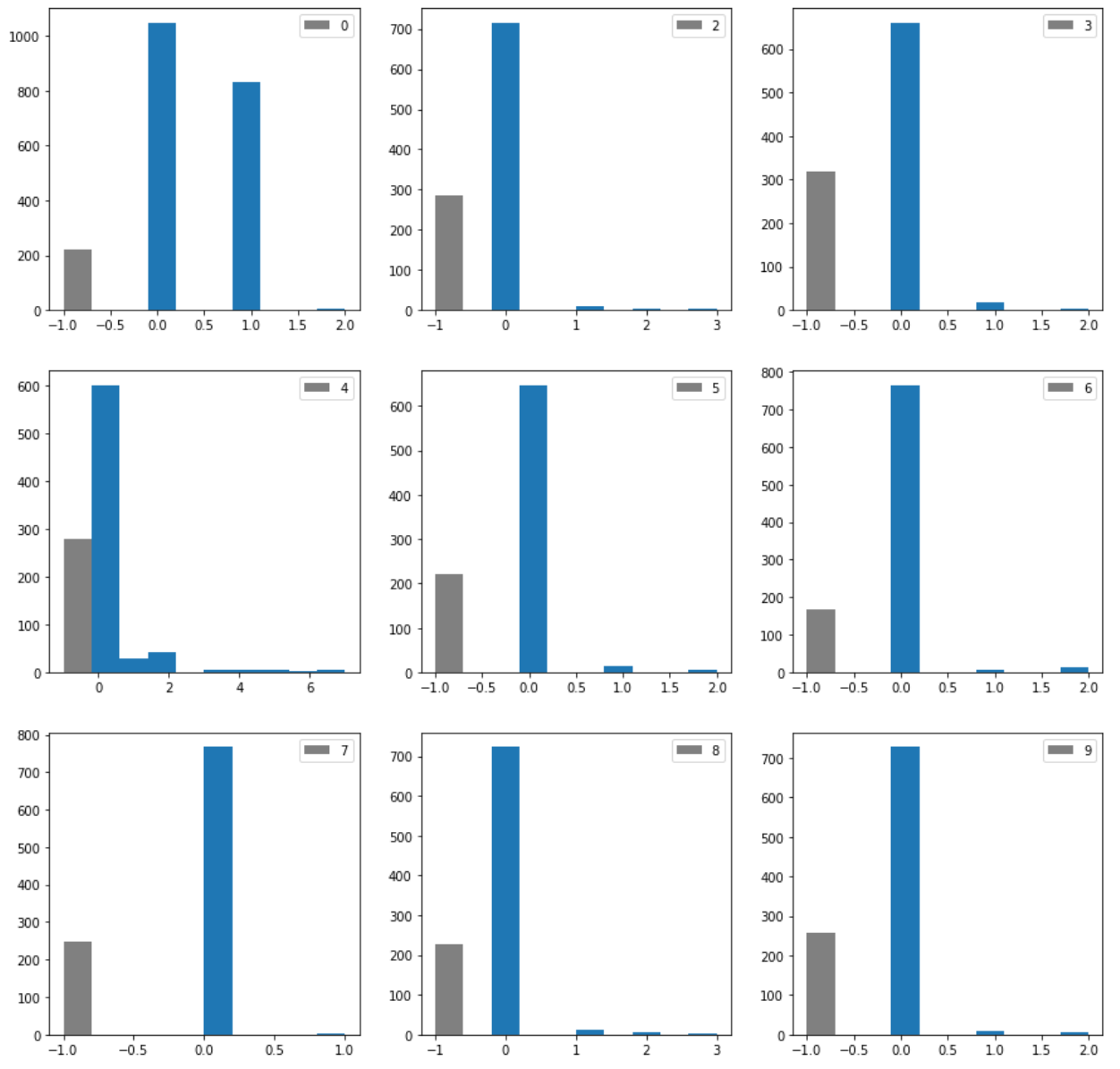}
    \caption{Intraclass fragmentation histogram viewed with DBSCAN on bridged
    MNIST data (bridged labels 0 and 1). As before, the \(x\) axis is the class
    label and the \(y\) axis is the count of instances in that label. Note that
    in the bridged case \(0\to 1\) there are two main bars (+ a gray noise bar)
    and in the rest there is only one main bar.}\label{fig:mnist-keras-shap-bridged-0-1-all}
\end{figure}

\begin{figure}[!t]
    \centering
    \includegraphics[width=\linewidth]{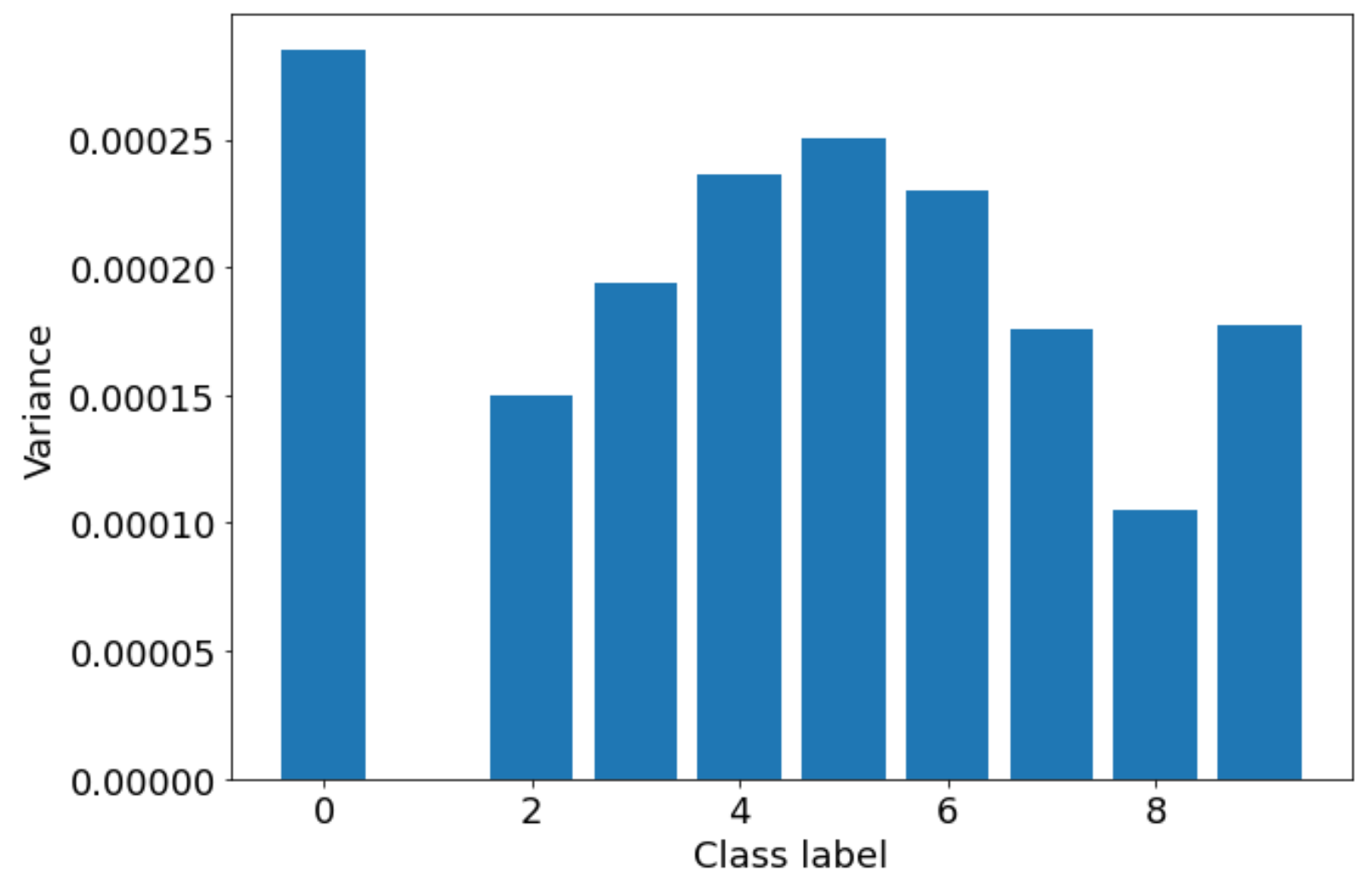}
    \caption{Variance inside classes in bridged MNIST data (bridged labels 0
    and 1). Note that the variance is highest in class \(0\) (the bridged
    class).}\label{fig:mnist-keras-shap-bridged-0-1-variance}
\end{figure}

In this case, PCA was used to reduce the dimensionality from \(784\) to \(5\)
as a preprocessing step, yielding the principal components seen in
Figure~\ref{fig:mnist-keras-shap-bridged-pca}. PCA was calculated from the set
of all explanations for instances the network predicted, giving a global set of
basis vectors. Running DBSCAN over this data with \(\epsilon = 0.004\) yielded
two distinct classes.  As a control, all other classes were calculated using
the same method, and this yielded the class membership visible in
Figure~\ref{fig:mnist-keras-shap-bridged-0-1-all}. This shows that it is
possible to find latent structure with this method. Additionally, even without
DBSCAN the latent class is visible in the variance explained by the PCA
vectors, as seen in Figure~\ref{fig:mnist-keras-shap-bridged-0-1-variance}.

To verify that this was not a fluke, this process was repeated for all possible
pairings (\(45\) in total). To facilitate this computation, a pipeline was
created to automatically run all cases, and a web application was produced in
order to view results. The code for both of these are available on
Github\footnote{\url{https://github.com/pkage/class_introspection_krhcai}}.

\section{Discussion}\label{chapter:results-limitations}

\subsection{Comparison to Baseline}

Compared to the baseline method described in
Section~\ref{section:meth-baseline}, the SHAP + PCA + DBSCAN method described
above is very effective at determining the latent structure.  This variation in
performance is indicative of the difference in the type of data being
processed. The baseline operates directly over the glyphs themselves, while the
class introspection pipeline operates over the classifier's
\textit{explanations} of the glyphs. While similar, the distinction is
important: in the baseline, the precise positions of the pixels are salient to
the class representation, whilst in the explanations it is the specific neuron
firings (and their intensities) that are salient. This allows the class
introspection pipeline to group positive classifications by the specific
neurons that are firing; the intuition being that the exact structure of the
glyph does not matter so long as the specific neurons with that class are
active which is fine tuned already by the classifier. Compare this to the
baseline, which is relegated to determining the specific pixels that make up a
class without the benefit of a trained neural network behind it.

\subsection{Limitations}


This experiment has shown that class introspection is a viable technique, but
there are still several limitations in its current iteration. One main issue is
that the explainability methods are imperfect. In most cases, SHAP or LIME
produce saliencies that are reasonable, but in others they pick up on specific
pixels that are assigned a saliency much higher than it's neighbors---or in
fact, any other pixel in any other instance---by several orders of magnitude.
This is infrequent, but care must be taken to avoid these outliers skewing the
PCA vectors.

Another issue is the dependence on the DBSCAN hyperparameter \(\varepsilon\).
Without careful tuning, this can very easily yield a "collapsed solution",
where the entire class is lumped into a single class, or a largely noise-filled
clustering solution where most points are in the noise class or in
minimum-cluster-size clusters. Future work here should be in adding checks to
ensure that the class distribution is meaningful.

In a similar vein, PCA is ill-suited to image data. PCA does not preserve the
structure of the image, rather flattening it out into a single-dimensional
vector. This flattening ruins any sort of spatial correlation between features,
and is extremely fragile to image scaling, rotation,
transposition, etc. A better choice for future work is a proper
computer-vision-based feature extraction method, such as Scale-Invariant
Feature Transform (SIFT) or Histogram of Oriented Gradients (HOG). These
techniques would preserve the feature positions in the saliency map, and allow
for a higher-quality dimensionality reduction. In its current iteration, class
introspection is ill-suited to complex image data (such as CIFAR or ImageNet)
due to this limitation with PCA.

Fundamentally, class introspection is limited by the characteristics (and
statistical power) of the classifier being explained. If the classifier does
not learn the differentiating features of a class with latent subclasses and
instead just treats that class as a catch-all, then class introspection is less
effective at discovering that subclass. This is why the neural-network-based
approaches were effective: the MNIST data was complex enough that the neural
network had to actually understand the input features, and this allowed the
bridged class to be easily discovered.

More fundamentally still, class introspection is an ``unknown-unknown
problem'': namely, the number of latent classes is unknown---and even worse,
the very existence of those latent classes is unknown. It is not unlike
searching for a needle in a haystack without even knowing if the needle is
there. Additionally, there is another problem: of these discovered subclasses,
there is no way of knowing which ones of those subclasses are intentional or
which are novel. This means that there will always have to be a human in the
loop to determine which subclasses are relevant.

\subsection{Future Work}\label{section:future-work}



The class introspection pipeline as it stands performs well in this paper, but
there are areas that can be improved for more robust performance on a wider
variety of data. A potential area of exploration is replacing the clustering
algorithm with a Gaussian mixture-based approach, where each point could be
maximized for the probability of being assigned a latent class instead of a
definite class label being assigned to each point. This could help identify
classes where the presence of latent structure is uncertain, but not definite.

Another direction of exploration is the even more involved technique of
comparing not only the input saliency but the neuron activations chain through
the last layers of the network for any particular instance. This would limit the
technique to only neural networks, but may be a powerful tool for identifying
subclasses as the last layers of the network can be effectively used for determining
similarity of instances \cite{paterson_detection_2019,caron2019deep}.

This technique has far reaching possibilities, as the ability to reliably
discover latent classes has implications across diverse fields of data science
and artificial intelligence research. For example, the ability to reliably detect
latent structure in classifiers greatly benefits medical studies who may be
looking for a robust alternative to subgroup analysis. More fundamentally,
this technique may allow for the auditing of classifiers to ensure that the
data they are trained on and the data that they work over is free of unknown
and potentially unwanted subclasses that could reduce the overall effectiveness
of the classifier.


\subsection{Conclusion}

Over the course of this project a technique for reliably extracting unlabeled
(latent) subclasses from existing datasets has been developed, leveraging
explainability techniques to take advantage of the statistical power of complex
models. This provided an advantage over other approaches for capturing latent
structure, and was demonstrated over an example dataset with
artificially-induced latent structure. 


\bibliographystyle{kr}
\bibliography{citations_union}

\begin{thebibliography}{}

\bibitem[\protect\citeauthoryear{Abdalla}{2021}]{abdalla_visual_2021}
Abdalla, A.
\newblock 2021.
\newblock A {{Visual History}} of {{Interpretation}} for {{Image Recognition}}.
\newblock {\em The Gradient}.

\bibitem[\protect\citeauthoryear{Asano, Rupprecht, and
  Vedaldi}{2020}]{asano2020selflabelling}
Asano, Y.~M.; Rupprecht, C.; and Vedaldi, A.
\newblock 2020.
\newblock Self-labelling via simultaneous clustering and representation
  learning.

\bibitem[\protect\citeauthoryear{Bell}{2021}]{bell_automatic_2021}
Bell, P.
\newblock 2021.
\newblock Automatic speech recognition course notes.

\bibitem[\protect\citeauthoryear{Caron \bgroup et al\mbox.\egroup
  }{2019}]{caron2019deep}
Caron, M.; Bojanowski, P.; Joulin, A.; and Douze, M.
\newblock 2019.
\newblock Deep clustering for unsupervised learning of visual features.

\bibitem[\protect\citeauthoryear{Do{\v s}ilovi{\'c}, Br{\v c}i{\'c}, and
  Hlupi{\'c}}{2018}]{dosilovic_explainable_2018}
Do{\v s}ilovi{\'c}, F.~K.; Br{\v c}i{\'c}, M.; and Hlupi{\'c}, N.
\newblock 2018.
\newblock Explainable artificial intelligence: {{A}} survey.
\newblock In {\em 2018 41st {{International Convention}} on {{Information}} and
  {{Communication Technology}}, {{Electronics}} and {{Microelectronics}}
  ({{MIPRO}})},  0210--0215.

\bibitem[\protect\citeauthoryear{Ester \bgroup et al\mbox.\egroup
  }{1996}]{ester_density-based_1996}
Ester, M.; Kriegel, H.-P.; Sander, J.; and Xu, X.
\newblock 1996.
\newblock A density-based algorithm for discovering clusters in large spatial
  databases with noise.
\newblock  226--231.
\newblock {AAAI Press}.

\bibitem[\protect\citeauthoryear{Ghai \bgroup et al\mbox.\egroup
  }{2020}]{ghai_explainable_2020}
Ghai, B.; Liao, Q.~V.; Zhang, Y.; Bellamy, R.; and Mueller, K.
\newblock 2020.
\newblock Explainable {{Active Learning}} ({{XAL}}): {{An Empirical Study}} of
  {{How Local Explanations Impact Annotator Experience}}.
\newblock {\em arXiv:2001.09219 [cs]}.

\bibitem[\protect\citeauthoryear{Higgins and
  Green}{2011}]{higgins_cochrane_2011}
Higgins, J.~P., and Green, S.
\newblock 2011.
\newblock Cochrane {{Handbook}} for {{Systematic Reviews}} of
  {{Interventions}}.
\newblock
  https://handbook-5-1.cochrane.org/chapter\_9/9\_6\_2\_what\_are\_subgroup\_analyses.htm.

\bibitem[\protect\citeauthoryear{Lanza and Rhoades}{2013}]{lanza_latent_2013}
Lanza, S.~T., and Rhoades, B.~L.
\newblock 2013.
\newblock Latent {{Class Analysis}}: {{An Alternative Perspective}} on
  {{Subgroup Analysis}} in {{Prevention}} and {{Treatment}}.
\newblock {\em Prevention science : the official journal of the Society for
  Prevention Research} 14(2):157--168.

\bibitem[\protect\citeauthoryear{LeCun and Cortes}{2010}]{lecun_mnist_2010}
LeCun, Y., and Cortes, C.
\newblock 2010.
\newblock {{MNIST}} handwritten digit database.

\bibitem[\protect\citeauthoryear{Lundberg and
  Lee}{2017}]{lundberg_unified_2017}
Lundberg, S.~M., and Lee, S.-I.
\newblock 2017.
\newblock A {{Unified Approach}} to {{Interpreting Model Predictions}}.
\newblock In Guyon, I.; Luxburg, U.~V.; Bengio, S.; Wallach, H.; Fergus, R.;
  Vishwanathan, S.; and Garnett, R., eds., {\em Advances in {{Neural
  Information Processing Systems}}}, volume~30,  4765--4774.
\newblock {Curran Associates, Inc.}

\bibitem[\protect\citeauthoryear{Lundberg}{2021}]{slundberg_shap_2021}
Lundberg, S.
\newblock 2021.
\newblock Slundberg/shap.

\bibitem[\protect\citeauthoryear{Marimont and
  Shapiro}{1979}]{marimont_nearest_1979}
Marimont, R.~B., and Shapiro, M.~B.
\newblock 1979.
\newblock Nearest {{Neighbour Searches}} and the {{Curse}} of
  {{Dimensionality}}.
\newblock {\em IMA Journal of Applied Mathematics} 24(1):59--70.

\bibitem[\protect\citeauthoryear{Paterson and
  Calinescu}{2019}]{paterson_detection_2019}
Paterson, C., and Calinescu, R.
\newblock 2019.
\newblock Detection and {{Mitigation}} of {{Rare Subclasses}} in {{Neural
  Network Classifiers}}.
\newblock {\em arXiv:1911.12780 [cs, stat]}.

\bibitem[\protect\citeauthoryear{Ribeiro, Singh, and
  Guestrin}{2016}]{ribeiro_why_2016}
Ribeiro, M.~T.; Singh, S.; and Guestrin, C.
\newblock 2016.
\newblock "{{Why Should I Trust You}}?": {{Explaining}} the {{Predictions}} of
  {{Any Classifier}}.
\newblock {\em arXiv:1602.04938 [cs, stat]}.

\bibitem[\protect\citeauthoryear{Scarpa}{2011}]{scarpa_data_2011}
Scarpa, B.
\newblock 2011.
\newblock Data {{Mining}}.
\newblock In Lovric, M., ed., {\em International {{Encyclopedia}} of
  {{Statistical Science}}}. {Berlin, Heidelberg}: {Springer}.
\newblock  336--339.

\bibitem[\protect\citeauthoryear{Shrikumar, Greenside, and
  Kundaje}{2019}]{shrikumar_learning_2019}
Shrikumar, A.; Greenside, P.; and Kundaje, A.
\newblock 2019.
\newblock Learning {{Important Features Through Propagating Activation
  Differences}}.
\newblock {\em arXiv:1704.02685 [cs]}.

\bibitem[\protect\citeauthoryear{Siedel}{2011}]{siedel_mixture_2011}
Siedel, W.
\newblock 2011.
\newblock Mixture {{Models}}.
\newblock In Lovric, M., ed., {\em International {{Encyclopedia}} of
  {{Statistical Science}}}. {Berlin, Heidelberg}: {Springer}.
\newblock  827--829.

\bibitem[\protect\citeauthoryear{Simonyan, Vedaldi, and
  Zisserman}{2014}]{simonyan_deep_2014}
Simonyan, K.; Vedaldi, A.; and Zisserman, A.
\newblock 2014.
\newblock Deep {{Inside Convolutional Networks}}: {{Visualising Image
  Classification Models}} and {{Saliency Maps}}.
\newblock {\em arXiv:1312.6034 [cs]}.

\bibitem[\protect\citeauthoryear{Teso and Kersting}{2018}]{teso_why_2018}
Teso, S., and Kersting, K.
\newblock 2018.
\newblock "{{Why Should I Trust Interactive Learners}}?" {{Explaining
  Interactive Queries}} of {{Classifiers}} to {{Users}}.
\newblock {\em arXiv:1805.08578 [cs, stat]}.

\bibitem[\protect\citeauthoryear{Teso}{2019}]{teso_toward_2019}
Teso, S.
\newblock 2019.
\newblock Toward {{Faithful Explanatory Active Learning}} with
  {{Self}}-explainable {{Neural Nets}}.
\newblock ~13.

\end{thebibliography}

\end{document}